\def\version{LNCS}
\newif\ifsubmission   \submissionfalse
\newif\ifUSENIX       \USENIXfalse
\newif\ifLNCS         \LNCSfalse
\newif\ifACM          \ACMfalse
\newif\iffull         \fullfalse

\newif \iffull 
\newif \ifACM
\newif \ifUSENIX
\newif \ifIEEE
\newif \ifLNCS

\newif \ifCCS
\newif \ifSP
\newif \ifNDSS
\newif \ifCrypto
\newif \ifFC

\def\fullstring{full}
\def\ACMstring{ACM}
\def\USENIXstring{USENIX}
\def\IEEEstring{IEEE}
\def\LNCSstring{LNCS}

\def\CCSstring{CCS}
\def\SPstring{SP}
\def\NDSSstring{NDSS}
\def\Cryptostring{CRYPTO}
\def\FCstring{FC}

\ifx \version\fullstring \fulltrue \fi
\ifx \version\ACMstring \ACMtrue \fi
\ifx \version\USENIXstring \USENIXtrue \fi
\ifx \version\IEEEstring \IEEEtrue \fi
\ifx \version\LNCSstring \LNCStrue \fi

\ifx \version\CCSstring \ACMtrue \CCStrue \fi
\ifx \version\SPstring \IEEEtrue \SPtrue \fi
\ifx \version\NDSSstring \IEEEtrue \NDSStrue \fi
\ifx \version\Cryptostring \LNCStrue \Cryptotrue \fi
\ifx \version\FCstring \LNCStrue \FCtrue \fi

\iffull \input{Conferences/FullVersion/fullversion.tex} \fi
\ifACM \input{Conferences/ACM/acm.tex} \fi
\ifUSENIX \input{Conferences/USENIX/usenix.tex} \fi
\ifIEEE \input{Conferences/IEEE/IEEE.tex} \fi
\ifLNCS 
\PassOptionsToPackage{dvipsnames}{xcolor}
\documentclass[runningheads]{Conferences/LNCS/llncs}

\ifsubmission \author{} \institute{} \fi
 \fi

\iffull \bibliography{references} \fi
\newif \ifcomments \commentsfalse
\newif \ifanon

\ifUSENIX \else \usepackage[table]{xcolor} \fi
\usepackage{xurl}
\usepackage{tabularx}
\usepackage{makecell}
\usepackage{booktabs}
\usepackage{xspace}
\usepackage{float}
\usepackage{placeins}
\usepackage{hyperref}
\hypersetup{
    colorlinks=true,
    allcolors=blue,
    linkcolor=red,
    filecolor=magenta,
}
\usepackage[many]{tcolorbox}
\usepackage{cleveref}
\usepackage{enumitem}
\setlist{nosep, topsep=1pt, leftmargin=*}
\usepackage{multirow}
\usepackage[figuresright]{rotating}
\usepackage{array}
\usepackage{nicematrix}
\usepackage{tikz}
\usetikzlibrary{shapes, arrows.meta, positioning, calc, math, tikzmark, fit}

\usepackage{amsfonts,amsmath}
\ifLNCS\else\usepackage{amsthm}\fi
\ifACM\else\usepackage{amssymb}\fi

\definecolor{ForestGreen}{RGB}{34,139,34}

\ifcomments
    \newcommand{\jay}[1]{\textsf{\small{\color{red}{[Jay: {#1}]}}}}
    \newcommand{\amy}[1]{\textsf{\small{\color{orange}{[Amy: {#1}]}}}}
    \newcommand{\danning}[1]{\textsf{\small{\color{blue}{[Danning: {#1}]}}}}
    \newcommand{\ari}[1]{\textsf{\small{\color{red}{[Ari: {#1}]}}}}
    \newcommand{\andres}[1]{\textsf{\small{\color{red}{[Andres: {#1}]}}}}
    \newcommand{\james}[1]{\textsf{\small{\color{green!90!black}{[James: {#1}]}}}}
\else
    \newcommand{\jay}[1]{}
    \newcommand{\amy}[1]{}
    \newcommand{\danning}[1]{}
    \newcommand{\ari}[1]{}
    \newcommand{\andres}[1]{}
    \newcommand{\james}[1]{}
\fi

\tcbset{
    sharp corners,
    colback = white,
    before skip = 0.1cm,
    after skip = 0.2cm
}                           

\newtcolorbox{boxA}{
    fontupper = \bf,
    boxrule = 1.5pt,
    colframe = black 
}

\iffull \newcommand{\mypara}{\paragraph}
\else \newcommand{\mypara}[1]{\vspace{2pt}\noindent\textbf{#1}\;} \fi

\iffull 
\else  \fi

\ifLNCS\else

\iffull \newtheorem{theorem}{Theorem}[section]
\else  \fi

\theoremstyle{definition}

\theoremstyle{remark}

\fi

\title{Paper Agents, Paper Gains: An Empirical Analysis of DeFi Investment Agents}
\titlerunning{Paper Agents, Paper Gains}

\ifanon
  \author{}
  \authorrunning{Anonymous}
  \institute{}
\else
  \author{Jay Yu\inst{1,2} \and Amy Zhao\inst{3,4} \and Danning Sui\inst{1}}
  \authorrunning{J. Yu et al.}
  \institute{
    Pantera Capital \and
    Stanford University \and
    IC3 \and
    Ava Labs
  }
\fi

\usepackage{algorithm}
\usepackage{algpseudocode}

\begin{document}
\maketitle

\begin{abstract}
DeFi investment agents---systems that attempt to use AI for autonomous on-chain trading---have attained over \$3 billion in combined token valuations since late 2024. We survey over 1,900 AI-tagged crypto projects, filter to investment-focused agents, and curate a set of 10 representative projects spanning strategy and observability dimensions. We then conduct a deep-dive architectural analysis of two prominent agent frameworks---ElizaOS (an open-source trading-agent runtime) and Virtuals Protocol (a launchpad for creating and tokenizing agents)---and a quantitative on-chain performance analysis of 11 Solana-based agent treasuries with publicly attributable trading activity, covering 925,323 token holders. We find that current deployments remain early and heterogeneous: (1) in our sampled projects, many do not yet provide clear evidence of autonomous trade execution, and developer interviews suggest that a substantial share of visible current deployments remain basic API integrations; (2) agent treasuries retain over \$30M in paper gains while their token holders collectively lost \$191.7M, with the top 1\% of wallets capturing 81.4\% of all gains, or \$1.81B; (3) token valuations are weakly connected to treasury fundamentals, with market-cap-to-AUM ratios exceeding 10,000$\times$ versus below 1$\times$ for established DeFi protocols; and (4) aggregate user gains peaked at \$2.4B before declining to net losses, with median returns negative on every platform and tokens declining 93\% on average from all-time highs. We interpret these outcomes as characteristic of a permissionless, first-generation market in which open platform infrastructure enables rapid experimentation but also allows naive or speculative agents to launch before robust standards for autonomy, performance, and stakeholder alignment have emerged. We therefore propose a maturity framework along three dimensions---autonomous execution, risk-adjusted profitability, and stakeholder alignment---to characterize the gap between current deployments and future investment-grade agent systems.

\end{abstract}

\section{Introduction}
\label{sec:intro}

Since late 2024, a wave of projects has emerged that present autonomous AI agents as a way to manage pooled capital in decentralized finance (DeFi) markets. These \emph{DeFi investment agents} span a range of ambitions: some, such as Giza~\cite{proj:GizaProtocol}, pursue passive strategies like yield optimization and vault rebalancing; others, such as AIXBT~\cite{proj:AIXBT}, provide active investment signals derived from social media sentiment and on-chain analytics; and projects like Eliza (formerly ai16z)~\cite{crypto_elizaos} attempt to build fully autonomous trading agents that execute on-chain transactions with capital bootstrapped through a publicly traded token. These projects have attracted significant liquidity in the cryptocurrency markets, with peak valuations exceeding \$3 billion \cite{coinbase_ai16z}. This emergence was anchored by two events: Virtuals Protocol's launch of bonding-curve agent token issuance in October 2024 and the rapid market emergence of ai16z/Eliza in November 2024. Combined AI-agent token market capitalizations peaked in January 2025 before contracting 88--99\% through mid-2025, tracking the broader pullback in AI-narrative and memecoin tokens over the same window.

Despite this rapid growth, several fundamental properties of DeFi investment agents remain empirically underexplored, including the degree to which trading decisions are executed algorithmically rather than manually, the profitability of the systems, transparency of the algorithms \cite{fabrega2026coinalg}, and the alignment between treasury-level performance and the returns realized by token holders. Our paper provides a systematic investigation of these properties, presenting a qualitative framework to assess the scope of DeFi investment agents, an analysis of two notable agent frameworks---ElizaOS (an open-source trading-agent runtime) and Virtuals Protocol (a launchpad for creating and tokenizing agents)---and an empirical evaluation of various on-chain trading performance for the agents built on these frameworks.

\subsection{Research Questions}
Our study addresses four key questions:
\begin{enumerate}
\item \textbf{Scope}: What is the broad landscape of DeFi investment agents currently deployed, and what strategies, design patterns, and adoption scales characterize these agents?
\item \textbf{Frameworks}: Several DeFi investment agents are built on shared open-source platforms---agent frameworks---that provide common infrastructure for wallet management, strategy execution, and token launch mechanics. How do the two leading frameworks, ElizaOS and Virtuals Protocol, structure agent design and execution, and what trading behaviors emerge from agents built on them?
\item \textbf{Performance}: What does publicly observable on-chain data reveal about the trading performance and returns of active DeFi investment agents, both at the treasury level and from the perspective of token holders?
\item \textbf{Evaluation}: What criteria distinguish a DeFi investment agent that functions as a legitimate autonomous trading system from one that functions primarily as a speculative vehicle, and how do existing projects measure against them?
\end{enumerate}

\subsection{Contributions}
We adopt a mixed-methods approach combining on-chain data analysis with semi-structured developer interviews, consistent with prior survey methodologies for DAOs~\cite{Sharma2024DAOs}. Our study covers the period from October 2024 to November 2025. Specifically, we make the following contributions:
\begin{itemize}
    \item Starting from a broad candidate pool drawn from CoinGecko, a token-market aggregator, and Messari, a venture fundraising database, we define inclusion criteria for DeFi investment agents and select 10 representative projects, characterizing their strategy types, trade observability---the degree to which an agent's trades are visible on-chain, from low (off-chain APIs, no public agent wallets) to high (public wallet addresses enabling audits)---funding mechanisms, and adoption scale (Section~\ref{broad-landscape}).
    \item Through architecture analysis and developer interviews, we analyze the structure and agent ecosystems of ElizaOS and Virtuals Protocol, documenting design motivations, operational realities, and the possible gaps between envisioned autonomous capabilities and deployments that have emerged permissionlessly on the platform under the frameworks (Section~\ref{narrow-landscape}).
    \item We define and apply a quantitative framework for measuring DeFi investment agent profitability, analyzing 11 agent treasuries and 925,323 token holders, distinguishing between platform treasury performance and token-holder returns, and benchmarking against passive strategies (Section~\ref{performance}). This performance cohort partially overlaps with but is not identical to the 10-project broad survey above due to lack of universal onchain observability; selection criteria are detailed in Section~\ref{performance}.
    \item Based on our empirical findings, we propose a framework for evaluating DeFi investment agents along three dimensions: autonomous trading, risk-adjusted profitability exceeding passive benchmarks, and stakeholder value alignment between trading performance and token-holder returns (Section~\ref{sec:maturity}).
\end{itemize}

\section{Related Work}
\label{sec:related}

\textbf{Crypto Market Manipulation.} Xu and Livshits \cite{xu2019anatomy} anatomize cryptocurrency pump-and-dump schemes, documenting coordinated price inflation followed by rapid collapse and negative returns for late participants. The speculative dynamics we observe in DeFi investment agent tokens---synchronized peaks, 93\% average declines, and right-skewed user return distributions where early entrants profit at the expense of later participants---are structurally consistent with these patterns, though the mechanism differs: agent tokens are inflated by narrative hype around AI capabilities rather than by explicit coordination in messaging groups. Conlon et al.~\cite{Conlon2025MemecoinContagion} study contagion effects in memecoin markets, showing that price rallies propagate across correlated memecoin tokens regardless of underlying fundamentals, and that buyer enthusiasm during these rallies is driven by social-media-fueled speculation rather than informed valuation. Several of the agents we study funded their treasuries through memecoin launches, directly coupling agent token prices to these contagion-driven boom-and-bust cycles.

\textbf{DAO Governance.} Surveys of Decentralized Autonomous Organizations~\cite{Sharma2024DAOs} establish methodological precedent for mixed-methods analysis of on-chain governance systems, combining quantitative analysis of treasury flows, token distributions, and voting records with qualitative developer interviews. We adopt a similar approach applied to investment-focused agents rather than governance DAOs. Several of the agents we study (notably Eliza/ai16z) adopt DAO-like governance structures for treasury management, blurring the boundary between these categories.

\textbf{Algorithmic and AI-Driven Trading.} Research on algorithmic and high-frequency trading~\cite{ji2017robots} examines automated execution in regulated markets with established oversight mechanisms. More recently, LLM-based trading agents have attracted growing academic attention. Yang et al.~\cite{yang2023fingpt} introduce FinGPT, an open-source financial LLM with applications in robo-advising and algorithmic trading. A wave of concurrent benchmarking work evaluates whether LLM agents can trade profitably: Chen et al.~\cite{chen2025stockbench} find that most LLM agents fail to outperform a buy-and-hold baseline on DJIA stocks over a four-month horizon; Fan et al.~\cite{fan2025aitrader} extend this evaluation to U.S.\ stocks, A-shares, and cryptocurrencies, concluding that general intelligence does not translate to trading capability; and Qian et al.~\cite{qian2025agentstrade} show that system-level design choices---such as multi-agent collaboration, tool-use patterns, and memory mechanisms---exert a stronger influence on profitability than the choice of underlying LLM. In the cryptocurrency domain specifically, Luo et al.~\cite{luo2025llmcrypto} propose a multi-agent system for crypto portfolio management that outperforms single-agent baselines on the top 30 tokens by market capitalization. Collectively, these results suggest that LLMs remain unreliable autonomous traders---consistent with the developer assessment we document in Section~\ref{narrow-landscape}---while leaving open the question of whether agent framework design can compensate for individual model limitations. DeFi investment agents operate in a distinct environment from the settings these benchmarks evaluate: permissionless deployment, pseudonymous operators, no regulatory oversight, and the ability to bootstrap capital through public token issuance. This combination enables speculative dynamics---particularly the conflation of agent token valuation with trading performance---that would be constrained in traditional markets.

\textbf{Collective Investment Algorithms.} Most directly related is concurrent work on Collective Investment Algorithms (CoinAlgs)~\cite{fabrega2026coinalg}, which proves a fundamental tradeoff: CoinAlgs cannot simultaneously ensure economic fairness and avoid losing profit to arbitrage. The CoinAlg framework also shows that agents with publicly observable wallet addresses are structurally vulnerable to frontrunning~\cite{daian2020flash}, as adversaries can anticipate and exploit their on-chain trades. Where that work establishes theoretical impossibility results, our study provides an empirical characterization of deployed systems; we revisit the connection between our findings and the CoinAlg Bind in Section~\ref{sec:maturity}.

\section{Broad Landscape of DeFi Investment Agents} \label{broad-landscape}

To characterize the current landscape of DeFi investment agents, we survey the full universe of AI-branded crypto projects and narrow to a representative sample of investment-focused agents.

\subsection{Data Collection and Selection}

We drew from two complementary data sources: (1) CoinGecko~\cite{coingecko_ai_agents}, a token market data aggregator, from which we collected all AI-tagged projects with recorded trading volume between October 2024 and November 2025; and (2) Messari~\cite{messari_fundraising}, a venture fundraising database, from which we collected all AI-tagged entities with funding rounds closed between September 2024 and December 2025 (the wider window captures fundraising that precedes token launches). The combined dataset covers 1{,}900 projects with recorded token market capitalizations and 182 venture-funded entities totaling \$1.87 billion across 229 rounds, with 52 appearing in both sources.

The majority of these projects are infrastructure-oriented: AI compute networks (e.g., Render, Akash), data and oracle layers, model training platforms, and agent development frameworks. To isolate the subset relevant to our study, we define a \emph{DeFi investment agent} as a system satisfying three \emph{inclusion criteria}:

\begin{enumerate}
    \item \emph{DeFi application:} the system is deployed on permissionless blockchain protocols and interacts with DeFi primitives such as decentralized exchanges, lending platforms, or yield vaults. This excludes traditional roboadvisors (e.g., Wealthfront, Betterment) that may trade digital assets but operate through centralized intermediaries.
    \item \emph{Investment function:} the system's primary purpose is investment-related---directly managing capital, optimizing yield, or providing trading signals and investment advice---rather than providing infrastructure such as compute, data feeds, or developer tooling.
    \item \emph{AI-branded:} the system presents itself as using AI or machine learning (e.g., LLMs, reinforcement learning, sentiment analysis) to inform its investment process.
\end{enumerate}

Applying these three inclusion criteria yields 1{,}035 projects classifiable as DeFi-focused trading or investment agents. We further restrict to systems operated as publicly accessible services---typically with an associated token, managed treasury, or user-facing platform---excluding self-hosted trading bots (e.g., Hummingbot, Freqtrade). Boundaries are not always sharp: Numerai operates primarily off-chain with an on-chain staking component, and ElizaOS agents can be self-hosted, though all agents in our sample are operated by identifiable teams or DAOs. Criterion 3 is intentionally based on how projects \emph{present} themselves; evaluating the substance of these AI claims is a central goal of our study (Sections~\ref{narrow-landscape}--\ref{performance}).

From the filtered pool, we selected 10 projects (Table~\ref{tab:broad-landscape}) through purposive sampling, prioritizing the largest user bases or token market capitalizations within each strategy category (passive, advisory, active trading) while ensuring diversity across trade observability and funding mechanism. Our purposive sample is designed to span strategy and observability types, not to serve as an exhaustive census of trading-capable agents within each ecosystem. We do not claim statistical representativeness. We categorize each project along six dimensions (Table~\ref{tab:broad-landscape}): agent strategy, trade observability, source of funds, blockchain ecosystem, user base, and current AUM. Two dimensions merit definition:

\begin{itemize}
    \item \emph{Agent Strategy:} whether the agent engages in passive DeFi activities (vaults, staking), provides investment advice, or actively trades with pooled capital.
    \item \emph{Trade Observability:} whether an external observer can trace the agent's on-chain transactions. We classify this as \emph{low} (no public agent wallets), \emph{mid} (dashboards or social media but no verifiable on-chain addresses), or \emph{high} (public wallet addresses enabling transaction audits).
\end{itemize}

\subsection{Survey Results}

Table~\ref{tab:broad-landscape} presents the 10 selected projects. Three key patterns emerge from this survey.

\begin{table*}[!ht] 
  \centering
  \scriptsize
  \renewcommand{\arraystretch}{1}
  \begingroup
    \setlength{\tabcolsep}{5pt} 
    \begin{NiceTabular}{%
      >{\columncolor{white}\raggedright\arraybackslash}p{1.2cm}|
      >{\raggedright\arraybackslash}p{3.3cm}
      >{\raggedright\arraybackslash}p{.7cm}
      >{\raggedright\arraybackslash}p{1.4cm}
      >{\raggedright\arraybackslash}p{1.3cm}
      >{\raggedright\arraybackslash}p{0.9cm}
      >{\raggedright\arraybackslash}p{1cm}
    }[color-inside]
      \CodeBefore
        \rowcolors{2}{gray!20}{}
      \Body
      \toprule
      \rowcolor{gray!50}
      \textbf{Project} & \textbf{Agent Strategy} & \textbf{Observ -ability} &
      \textbf{Source of Funds} & \textbf{Chains} & \textbf{User Base} & \textbf{Current AUM (USD)} \\
      \midrule

      Axal \cite{proj:Axal} &
      Passive yield strategies - autonomous on-chain yield and vault strategies for non-technical users &
      Low &
      Individual user wallet &
      Base &
      -- &
      -- \\
      \midrule

      Giza \cite{proj:GizaProtocol} &
      Passive DeFi strategies - AI-driven allocation, arbitrage, and automated DeFi yield strategies &
      Low &
      Individual user wallet &
      Base &
      30{,}000 agents &
      6{,}888{,}486 \\
      \midrule

      TrueNorth \cite{proj:TrueNorthXYZ} &
      Active investment advice - agent generates trade suggestions and executes via user wallets &
      Low &
      Individual user wallet &
      Hyperliquid &
      2{,}000 beta testers &
      -- \\
      \midrule

      Surf \cite{proj:AskSurfAI} &
      Active investment advice - personal-agent service providing automated trade suggestions for crypto users &
      Low &
      Individual user wallet &
      -- &
      30{,}000 &
      -- \\
      \midrule

      Numerai Crypto \cite{proj:NumeraiCrypto} &
      Active trade simulation - hedge-fund–style agent simulating and aggregating crowd-sourced trading signals &
      Mid &
      NMR token and proprietary sources &
      -- &
      38{,}917 holders &
      -- \\
      \midrule

      AIXBT \cite{proj:AIXBT} &
      Active investment advice - public social agent tied to a memecoin with trade-suggestion logic &
      High &
      Launched own memecoin AIXBT &
      Base (Virtuals) &
      421{,}080 holders &
      3{,}393 \\
      \midrule

      Truth Terminal \cite{proj:TruthTerminal} &
      Active investment advice - public agent and social-signal trading model funded by GOAT memecoin &
      Mid &
      Proprietary and public memecoin GOAT &
      Solana &
      -- &
      17{,}485 \\
      \midrule

      Axelrod \cite{proj:AXR_AIXVC} &
      Active trading - public agent executing trades and yield strategies backed by the AXR token &
      High &
      AXR token &
      Base (Virtuals) &
      28{,}189 holders &
      -- \\
      \midrule

      ai16z / Eliza \cite{proj:ElizaOS} &
      Active trading - autonomous multi-agent trading system executing strategies via ElizaOS &
      High &
      Funded by ai16z token &
      Solana (ElizaOS) &
      101{,}025 holders &
      138{,}957 \\
      \midrule

      Alpha Arena \cite{proj:NoF1AI} &
      Active trading - autonomous LLM agents executing trades in a live benchmark environment &
      High &
      Proprietary &
      Hyperliquid &
      -- &
      42{,}400 \\      

      \bottomrule
    \end{NiceTabular}
  \endgroup
  \caption{Broad landscape of DeFi investment agents, categorized by agent function, observability, funding source, framework, user base, and AUM (as of November 30, 2025). ``--'' indicates data not publicly available. Rows are deployed agents; the ElizaOS and Virtuals Protocol frameworks appear only as labels in the chain column.}
  \label{tab:broad-landscape}
\end{table*}

\textbf{The vast majority of ``agents'' do not trade autonomously.} Of the 10 projects, only 3 actively execute trades with pooled funds (Axelrod, Eliza, Alpha Arena). Two engage solely in passive strategies such as vault allocation and staking (Axal, Giza). The remaining 5 provide investment advice or trade simulations without autonomous execution. TrueNorth, for instance, generates trade suggestions and can facilitate execution through user wallets, but requires user approval rather than trading autonomously with its own treasury.

\textbf{Funding mechanism correlates with trade observability.} Projects that bootstrapped capital through public token launches (AIXBT, Axelrod, Eliza, Truth Terminal) all have mid-to-high observability: tokens create on-chain footprints and incentives for community scrutiny, with AIXBT and GOAT (Truth Terminal) exhibiting memecoin dynamics~\cite{Conlon2025MemecoinContagion}. Conversely, the 4 projects funded through individual user wallets (Axal, Giza, TrueNorth, Surf) have low observability, as agent activity is dispersed across private wallet instances with no public treasury to audit. The remaining 2 projects---Numerai (NMR utility token plus proprietary sources) and Alpha Arena (proprietary capital)---do not fit either pattern cleanly. Truth Terminal additionally received private investment from venture capitalist Marc Andreessen~\cite{Bellan2024TruthTerminal}.

\textbf{Observability does not imply verifiable autonomy.} Even among the 4 high-observability projects with public blockchain addresses (AIXBT, Axelrod, Eliza, Alpha Arena), we could not independently verify whether transaction execution was truly autonomous or required manual signing and human intervention. On-chain data reveals \emph{what} an agent did, but not \emph{who} or \emph{what} initiated the action. The 2 mid-observability projects (Numerai, Truth Terminal) provide dashboards or social media posts describing trades but no on-chain addresses to verify execution; for the remaining 4, we were unable to identify publicly traceable agent activity sufficient for independent verification.

\textbf{Deep-dive selection.} Most projects in our sample operate through user-specific wallets, proprietary pipelines, or passive vault allocations, limiting verifiable claims about autonomy or performance. ElizaOS and Virtuals Protocol are exceptions: both deploy public agent wallet instances enabling on-chain inspection, maintain publicly traded tokens with substantial market participation, and support comparatively large user bases, making them the most empirically tractable agent frameworks in DeFi. We analyze them in depth in Section~\ref{narrow-landscape}.
\section{ElizaOS and Virtuals Protocol Ecosystem Analysis} \label{narrow-landscape}

\subsection{Framework Architecture and Developer Perspectives}

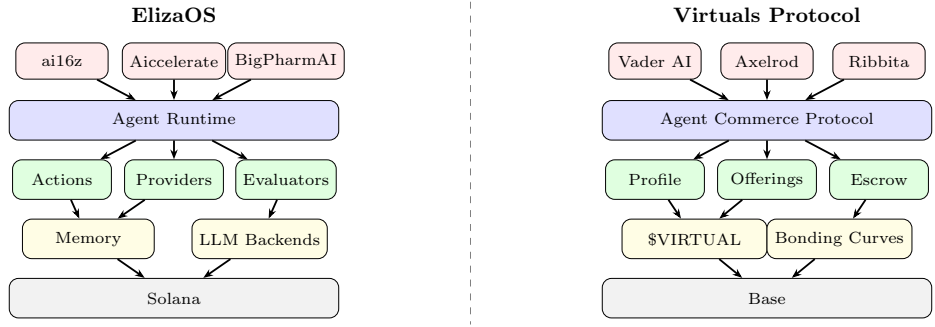
\begin{figure*}[t]
\centering
\resizebox{\textwidth}{!}{%
\begin{tikzpicture}[
    agent/.style={draw, rounded corners, minimum height=0.55cm, minimum width=1.4cm, font=\scriptsize, fill=white},
    box/.style={draw, rounded corners, minimum height=0.6cm, minimum width=2cm, font=\scriptsize, align=center},
    layer/.style={draw, rounded corners, minimum height=0.6cm, font=\scriptsize, fill=gray!10},
    arrow/.style={-{Stealth[length=1.5mm]}, thick},
    label/.style={font=\small\bfseries},
]

\node[label] at (-5.5, 3.1) {ElizaOS};

\node[agent, fill=red!8] (eliza) at (-7.2, 2.4) {ai16z};
\node[agent, fill=red!8] (aicc) at (-5.5, 2.4) {Aiccelerate};
\node[agent, fill=red!8] (pharma) at (-3.8, 2.4) {BigPharmAI};

\node[box, fill=blue!12, minimum width=5cm] (runtime) at (-5.5, 1.5) {Agent Runtime};

\node[box, fill=green!12, minimum width=1.5cm] (actions) at (-7.2, 0.6) {Actions};
\node[box, fill=green!12, minimum width=1.5cm] (providers) at (-5.5, 0.6) {Providers};
\node[box, fill=green!12, minimum width=1.5cm] (evaluators) at (-3.8, 0.6) {Evaluators};

\node[box, fill=yellow!12, minimum width=2cm] (memory) at (-6.8, -0.3) {Memory};
\node[box, fill=yellow!12, minimum width=2cm] (llm) at (-4.2, -0.3) {LLM Backends};

\node[layer, minimum width=5cm] (chain1) at (-5.5, -1.2) {Solana};

\draw[arrow] (eliza) -- (runtime);
\draw[arrow] (aicc) -- (runtime);
\draw[arrow] (pharma) -- (runtime);
\draw[arrow] (runtime) -- (actions);
\draw[arrow] (runtime) -- (providers);
\draw[arrow] (runtime) -- (evaluators);
\draw[arrow] (actions) -- (memory);
\draw[arrow] (providers) -- (memory);
\draw[arrow] (evaluators) -- (llm);
\draw[arrow] (memory) -- (chain1);
\draw[arrow] (llm) -- (chain1);

\node[label] at (3.5, 3.1) {Virtuals Protocol};

\node[agent, fill=red!8] (vader) at (1.8, 2.4) {Vader AI};
\node[agent, fill=red!8] (axel) at (3.5, 2.4) {Axelrod};
\node[agent, fill=red!8] (ribbita) at (5.2, 2.4) {Ribbita};

\node[box, fill=blue!12, minimum width=5cm] (acp) at (3.5, 1.5) {Agent Commerce Protocol};

\node[box, fill=green!12, minimum width=1.5cm] (profile) at (1.8, 0.6) {Profile};
\node[box, fill=green!12, minimum width=1.5cm] (offerings) at (3.5, 0.6) {Offerings};
\node[box, fill=green!12, minimum width=1.5cm] (escrow) at (5.2, 0.6) {Escrow};

\node[box, fill=yellow!12, minimum width=2.2cm] (token) at (2.4, -0.3) {\$VIRTUAL};
\node[box, fill=yellow!12, minimum width=2.2cm] (bonding) at (4.6, -0.3) {Bonding Curves};

\node[layer, minimum width=5cm] (chain2) at (3.5, -1.2) {Base};

\draw[arrow] (vader) -- (acp);
\draw[arrow] (axel) -- (acp);
\draw[arrow] (ribbita) -- (acp);
\draw[arrow] (acp) -- (profile);
\draw[arrow] (acp) -- (offerings);
\draw[arrow] (acp) -- (escrow);
\draw[arrow] (profile) -- (token);
\draw[arrow] (offerings) -- (token);
\draw[arrow] (escrow) -- (bonding);
\draw[arrow] (token) -- (chain2);
\draw[arrow] (bonding) -- (chain2);

\draw[dashed, gray] (-1, 3.3) -- (-1, -1.6);

\end{tikzpicture}%
}
\caption{Architecture of ElizaOS and Virtuals Protocol with representative deployed agents. Deployed agents (top row: ai16z, Aiccelerate, BigPharmAI on ElizaOS; Vader AI, Axelrod, Ribbita on Virtuals) sit above their respective runtime and settlement layers. ElizaOS provides a modular plugin system (Actions, Providers, Evaluators) with developer-controlled LLM selection. Virtuals standardizes agent commerce through escrow-based settlement and bonding curve tokenization.}
\label{fig:architecture}
\end{figure*}

We focus on ElizaOS and Virtuals Protocol as the two dominant paradigms for DeFi agent deployment. Both expose public wallet addresses, enabling the on-chain observability required for empirical analysis (Figure~\ref{fig:architecture}). To complement our technical analysis, we conducted semi-structured interviews with core contributors of two major agent frameworks
. We recruited them via direct outreach after they emerged as the dominant ecosystems in our broad-landscape filtering (Section~\ref{broad-landscape}). Interviews covered (a) original design intent, (b) observed agent behaviors in production, and (c) the gap between envisioned autonomy and current capability. We treat these interviews as targeted qualitative depth complementing our quantitative findings, not as a representative sample.

%

\textbf{ElizaOS}~\cite{crypto_elizaos,walters2025elizaweb3friendlyai} is an open-source TypeScript agent framework first popularized by the ai16z DAO in October 2024 with a flagship trading agent built on the codebase~\cite{decrypt_ai16z_launch}. It was rebranded from \emph{ai16z} to \emph{ElizaOS} in January 2025 to disambiguate from the a16z venture firm~\cite{decrypt_elizaos_rename}. Hereafter \emph{ai16z / Eliza} (Table~\ref{tab:broad-landscape}) refers to the deployed agent, and \emph{ElizaOS} to the framework.

Architecturally, ElizaOS is built around a unified plugin model: each message flows through a three-stage pipeline~\cite{walters2025elizaweb3friendlyai,elizaos_docs} in which \emph{Providers} inject contextual data (wallet balances, price feeds, conversation history) into the LLM prompt, the LLM selects from \emph{Actions} (executable capabilities such as token swaps), and \emph{Evaluators} run post-hoc to update persistent memory. On-chain trading is not built into the core but supplied by optional, opt-in plugins drawn from a registry of over 90~\cite{elizaos_plugin_registry}---e.g., the Solana plugin's Jupiter swaps and SPL transfers~\cite{elizaos_plugin_solana} or Hyperliquid spot trading~\cite{elizaos_plugin_hyperliquid}---so not all deployed ElizaOS agents can execute on-chain transactions. Crucially, while the framework itself is open-source, the plugin configuration of individual deployed agents is not publicly auditable: an on-chain wallet reveals what transactions were executed, but not whether they were initiated autonomously by the runtime or manually signed by a human operator.

The team described the original goal as ``aggregating alpha''---translating social sentiment, on-chain data, and private information into actionable trading signals---an approach targeting small-cap memecoins where information asymmetries enable trading advantages before price discovery. Their response emphasized a key limitation for current state of art of the market: \emph{``LLMs cannot trade well''} without human insight and domain-specific inputs. The framework's ultimate vision is a ``trust marketplace'' where humans provide proprietary intelligence and agents return actionable insights.

%

\textbf{Virtuals Protocol}~\cite{VirtualsProtocolWhitepaper2025} provides managed infrastructure for agent deployment on Base. Its decision-making engine, the GAME (Generative Autonomous Multimodal Entities) framework~\cite{virtuals_game_framework}, decomposes agent behavior hierarchically: a high-level planner selects among specialized workers, each executing sequences of functions including on-chain transactions. Trading capabilities are provided through community-contributed plugins and a Python/TypeScript SDK~\cite{virtuals_game_sdk}, though as with ElizaOS, plugin selection is developer-controlled.

Each agent is minted as an ERC-6551 NFT on Base~\cite{virtuals_protocol_contracts}, serving as both the agent's on-chain identity and wallet address (a Token Bound Account). Agent tokens launch on a bonding curve denominated in \$VIRTUAL and graduate to a Uniswap V2 liquidity pool upon reaching a capitalization threshold. An Agent Commerce Protocol (ACP)~\cite{virtuals_acp_deepdive} standardizes inter-agent service discovery and escrow-based settlement. Over 17,000 agents have launched on Virtuals as of September 2025.

The team identified two factors distinguishing successful agents: \emph{founder credibility} (developers who publicly identify themselves signal commitment) and \emph{domain expertise} in trading---``actually successful agents have required domain expertise,'' purely technical teams without trading experience struggle to build effective strategies. The survey characterized the technical maturity across the platform as limited: most agents implement basic API integrations rather than autonomous execution, with ``not a ton that are advanced execution level.''

Both interviews converged on common themes: trading domain expertise is essential for performance, trust remains a key barrier to adoption, and genuine autonomous execution is rare in current deployments. Taken together, the interviews suggest that verifiable advanced autonomous execution remains limited among the deployments most visible to our respondents. We revisit how these characterizations align with the observed user-PnL distribution in Section~\ref{performance}.

\subsection{Data Collection and Agent Selection}

We analyze agents with publicly observable trading activity deployed on ElizaOS (Solana) and Virtuals Protocol (Base) between October 2024 and November 2025. Agent selection required: (1) a publicly known treasury address, (2) an associated token with trading history, and (3) observable on-chain trading activity. Treasury addresses were identified through project documentation, official announcements, and block explorer verification. We computed market capitalization from DEX trading data, AUM from treasury wallet balances at daily snapshots, and user PnL from realized gains/losses of token holders who both purchased and sold during the study period.

Table~\ref{tab:narrow-landscape} summarizes 10 agents meeting these criteria, reporting peak market capitalization, ATH date, peak AUM, and top portfolio holdings.


\begin{table*}[!ht]
  \hspace*{-2cm}
  \centering
  \scriptsize
  \renewcommand{\arraystretch}{1.20}
  \begingroup
    \setlength{\tabcolsep}{4pt}

    \newcommand{\solscanacct}[2]{\href{https://solscan.io/account/#1}{\texttt{#2}}}
    \newcommand{\basescan}[2]{\href{https://basescan.org/address/#1}{\texttt{#2}}}

    \begin{NiceTabular}{%
      >{\columncolor{white}\raggedright\arraybackslash}p{1.5cm}|
      >{\raggedright\arraybackslash}p{1.4cm}
      >{\raggedright\arraybackslash}p{1.4cm}
      >{\raggedright\arraybackslash}p{2cm}
      >{\raggedright\arraybackslash}p{1.35cm}
      >{\raggedright\arraybackslash}p{1.4cm}
      >{\raggedright\arraybackslash}p{1.6cm}
    }[color-inside]
      \CodeBefore
        \rowcolors{2}{gray!20}{}
      \Body
      \toprule
      \rowcolor{gray!50}
      \textbf{Project} & \textbf{Framework} & \textbf{Treasury Address} &
      \textbf{Peak Market Cap} & \textbf{ATH Date} & \textbf{Peak AUM} & \textbf{Major Holdings} \\
      \midrule

      ai16z / Eliza &
      ElizaOS &
      \solscanacct{AM84n1iLdxgVTAyENBcLdjXoyvjentTbu5Q6EpKV1PeG}{AM8…PeG} &
      \$2{,}716{,}989{,}656 & 1/9/2025 & \$22{,}963{,}939 & degenai, fxn, ai16z \\

      Aiccelerate DAO &
      ElizaOS &
      \solscanacct{CVe6u3rZnxpJkhCT8AZLW2nJG29rXSHHSoYzb8W1ffYf}{CVe…fYf} &
      \$274{,}998{,}502 & 1/11/2025 & \$171{,}956 & SOL, bAInance, Ropirito \\

      Vader AI &
      Virtuals &
      \basescan{0x731814e491571A2e9eE3c5b1F7f3b962eE8f4870}{0x7…870} &
      \$142{,}638{,}000 & 1/2/2025 & \$27{,}470 & ETH, USDC, VIRTUAL \\

      Big PharmAI &
      ElizaOS &
      \solscanacct{3WvMA1vKKPum73WSqeBhqu9pkvGnP2S1PS5Wsmyx56C5}{3Wv…6C5} &
      \$57{,}937{,}572 & 1/3/2025 & \$694{,}838 & BIO, SOL, Big PharmAI \\

      ICM &
      ElizaOS &
      \solscanacct{7HeqgxkGEmar88tFHNzhzPbFozRgncUS4CFJdmWaZTq6}{7He…Tq6} &
      \$28{,}360{,}906 & 1/6/2025 & \$152{,}874 & CPT, SPLAT, ICM \\

      Axelrod &
      Virtuals &
      \basescan{0x58Db197E91Bc8Cf1587F75850683e4bd0730e6BF}{0x5…6BF} &
      \$22{,}691{,}000 & 5/21/2025 & \$2{,}244 & VIRTUAL, ETH, cbBTC \\

      bAInance Labs &
      ElizaOS &
      \solscanacct{gGm7w2gxyWPMU5g5aAPSUZT4tzxDjpSB8FNw9Pfavd6}{gGm…vd6} &
      \$7{,}809{,}000 & 12/26/2024 & \$691{,}561 & WSOL, kotopia \\

      TOPKEK &
      ElizaOS &
      \solscanacct{BgwAen9wHPS5S93RngHQy6ZUHYsDQrZUm6BA76kjRyHC}{Bgw…yHC} &
      \$2{,}310{,}238 & 1/14/2025 & \$1{,}446{,}523 & SOL, Kolwaii, CRISPR \\

      kotopia &
      ElizaOS &
      \solscanacct{gsbJNUwQUYdCsRrmkKkkpJBYefKdwDp9rsLCZHJ1gMC}{gsb…gMC} &
      \$1{,}112{,}000 & 11/17/2024 & \$1{,}065{,}964 & SOL, USDC, koto \\

      Ribbita &
      Virtuals &
      \basescan{0xa4a2e2ca3fbfe21aed83471d28b6f65a233c6e00}{0xa4…e00} &
      \$428{,}160 & 10/27/2024 & \$25{,}317 & ETH, USDC, DEGEN \\

      \bottomrule
    \end{NiceTabular}
  \endgroup
  \caption{Peak market cap, AUM, and major token holdings for DeFi agents on ElizaOS and Virtuals Protocol. Holdings show the top three tokens by value in each treasury.}
  \label{tab:narrow-landscape}
\end{table*}

\textbf{Scale and Concentration.} While DeFi investment agents have achieved meaningful scale, with 3 of 10 agents having a peak market capitalization exceeding 100 million USD and 3 agents having a peak AUM of over 1 million USD, extreme concentration dominates both valuation and portfolio holdings, with Eliza having around 10 times the peak market capitalization and 20 times the peak AUM of the next largest agent. 

\textbf{Extreme Market Cap to AUM Ratios.} The ratio of token market capitalization to assets under management functions as an analog to equity price-to-book multiples, with elevated values reflecting speculative interest~\cite{qin2021empirical,qin2021attacking}. We find 8 of 10 agents have peak MC-to-AUM ratios exceeding 10, with Vader AI and Axelrod exceeding 10,000$\times$. This vastly exceeds industry benchmarks: major DeFi protocols (Maker, Aave, Curve, Uniswap) have MC-to-TVL ratios below 1~\cite{coinmarketcap}, and Digital Asset Treasuries---publicly-listed companies holding cryptocurrencies---trade at 1--3$\times$ NAV~\cite{PanteraDATboard}. These extreme ratios suggest valuations driven by narrative speculation rather than fundamentals.

\textbf{Timing.} Across these projects, we observe two timing regularities. First, peaks in market capitalization cluster tightly within a narrow window—primarily late 2024 to early 2025 for ElizaOS agents—indicating that valuation dynamics are driven less by idiosyncratic project events and more by shared narrative cycles within each framework. Second, for nearly all agents, peak AUM occurs \emph{after} the token’s ATH, often by several days or weeks, with Virtuals-based agents showing the longest lags. Together, these patterns suggest a reflexive process in which coordinated narrative waves generate synchronized price appreciation, and subsequent capital inflows into agent treasuries follow only once speculative activity has already peaked.

\textbf{Small-cap token portfolios.} The portfolio construction of surveyed DeFi investment agents skews heavily towards low-capitalization tokens. All projects in our study hold at least one token in their top portfolio positions with less than \$10 million USD market capitalization~\cite{CoinMarketCapSolanaEcosystem2025}, with most portfolios dominated by such small-cap assets alongside the native chain token (SOL or VIRTUAL). Unlike DAO treasuries such as Uniswap, Aave, and Arbitrum---whose top portfolio positions all exceed \$50 million USD in market capitalization---agent portfolios are dominated by tokens under \$10 million. This concentration in speculative assets suggests a preference for emerging or volatile tokens relative to established DAO treasuries, consistent with the ``memecoin'' focus described in our ElizaOS interview.

\section{Investment Performance and Holder Analysis} \label{performance}

We analyze the investment performance of DeFi agent platforms through on-chain treasury data and token holder returns. Our quantitative analysis focuses on 11 Solana-based AI agent platform projects: the 7 ElizaOS agents in Table~\ref{tab:narrow-landscape} plus 4 additional Solana-based platforms---Truth Terminal, Zerebro, Lola, and Nuclease DAO---that meet the same treasury observability criteria. We exclude Virtuals Protocol agents (deployed on Base) from this analysis because VIRTUAL serves as a shared routing currency across the platform---users must acquire VIRTUAL tokens before purchasing any agent token, making it difficult to attribute VIRTUAL holdings to participation in specific agents.

\subsection{PnL (Profit and Loss) Methodology}

\subsubsection{Data Sources and Forward-Filling}
Our analysis relies on two primary data sources from Dune Analytics~\cite{dune}: \texttt{solana\_utils.daily\_balances}, which records end-of-day token balances only on days when a wallet's balance changes, and \texttt{dex\_solana.price\_hour}, which provides hourly DEX prices aggregated to daily medians.

A key subtlety arises from Solana's account model: a single wallet can hold multiple token accounts for the same token mint, each tracked separately in the balance table. Consequently, for a given wallet-token pair on a single day, multiple records may exist corresponding to different token accounts. We address this with a two-stage preprocessing step: (1) for each wallet-account-token tuple, we forward-fill the most recent balance to construct a complete daily series per account; (2) we then aggregate across all accounts belonging to the same wallet to obtain the total daily balance per wallet-token pair.

Since both underlying tables are sparse (i.e., they do not contain entries for every calendar day), we construct complete daily time series through forward-filling: we create a daily grid and carry forward the most recent known balance and price to days with missing observations. This ensures consistent daily snapshots for PnL computation.

\subsubsection{FIFO Realized PnL}
We infer trades from daily balance changes: a net quantity increase $\Delta q_t > 0$ on day $t$ is treated as a ``buy'' at that day's median price $p_t$, while a net decrease $\Delta q_t < 0$ is treated as a ``sell.'' Realized PnL is computed using First-In, First-Out (FIFO) lot matching. For a sequence of $n$ buy lots with quantities $q_1, \ldots, q_n$ at prices $p_1, \ldots, p_n$, we track cumulative buy quantities $B_i = \sum_{j=1}^{i} q_j$. Similarly, for sell lots, we track cumulative sell quantities $S_i$. A sell lot $s$ with cumulative range $[S_{s-1}, S_s]$ matches buy lot $b$ with range $[B_{b-1}, B_b]$ for quantity:
\begin{equation}
m_{s,b} = \max\left(0, \min(B_b, S_s) - \max(B_{b-1}, S_{s-1})\right)
\end{equation}
The realized PnL for sell $s$ is then:
\begin{equation}
R_s = \sum_{b} m_{s,b} \cdot (p_s^{\text{sell}} - p_b^{\text{buy}})
\end{equation}
where $p_s^{\text{sell}}$ and $p_b^{\text{buy}}$ are the respective sell and buy prices. Cumulative realized PnL up to day $t$ is $R_t = \sum_{\tau \leq t} R_\tau$, which remains constant on days without sales.

\subsubsection{Mark-to-Market (MTM) Unrealized PnL}
For remaining (unsold) positions, we compute daily unrealized PnL as the difference between current market value and remaining cost basis. Let $C_t^{\text{buy}}$ denote cumulative buy notional (total cost of all purchases) and $C_t^{\text{sold}}$ denote cumulative FIFO cost of sold lots up to day $t$. The remaining cost basis is:
\begin{equation}
\text{CB}_t = \max\left(C_t^{\text{buy}} - C_t^{\text{sold}}, 0\right)
\end{equation}
The unrealized PnL on day $t$ is then:
\begin{equation}
U_t = q_t \cdot p_t - \text{CB}_t
\end{equation}
where $q_t$ is the forward-filled end-of-day quantity and $p_t$ is the forward-filled price.

\subsubsection{Total PnL}
Total PnL on day $t$ combines cumulative realized gains with the current unrealized snapshot:
\begin{equation}
\text{PnL}_t^{\text{total}} = R_t + U_t
\end{equation}

\paragraph{Caveats.} Our methodology uses daily balance snapshots rather than individual trade events. We observe only the net change in token holdings per day, not the sequence of transactions that produced it. This approximation has two implications: (1) non-trade events such as transfers, mints, and airdrops are indistinguishable from purchases, and (2) intraday trading activity is collapsed into a single net flow, which may introduce errors when a wallet both buys and sells the same token within a day. Thus, rapid buy-sell round trips within the same day are not captured separately; this mainly affects high-frequency wallets and is unlikely to dominate retail-holder outcomes. Additionally, DEX trades can produce extreme price outliers; we apply token-specific price caps to exclude anomalous observations from the daily median calculation~\cite{dune:platformpnl}.

\subsubsection{Platform Treasury PnL}
Platform Treasury PnL measures the investment performance of the project's managed funds. We identify each project's treasury wallet address and aggregate PnL across all tokens in its portfolio. The methodology above is applied per treasury-token pair, then summed across all holdings~\cite{dune:platformpnl}.

\subsubsection{User PnL}
User PnL quantifies returns from the perspective of participants who hold the platform's native token. We apply the same FIFO realized and MTM unrealized methodology to compute each user's total PnL. Due to Dune Analytics' computational constraints, we download user balance snapshots~\cite{dune:userbalance} and daily token prices~\cite{dune:tokenprice} separately, then compute PnL via a Python script using the same logic.

\subsection{Market Dynamics}

\begin{figure}[t]
    \centering
    \includegraphics[width=0.95\linewidth]{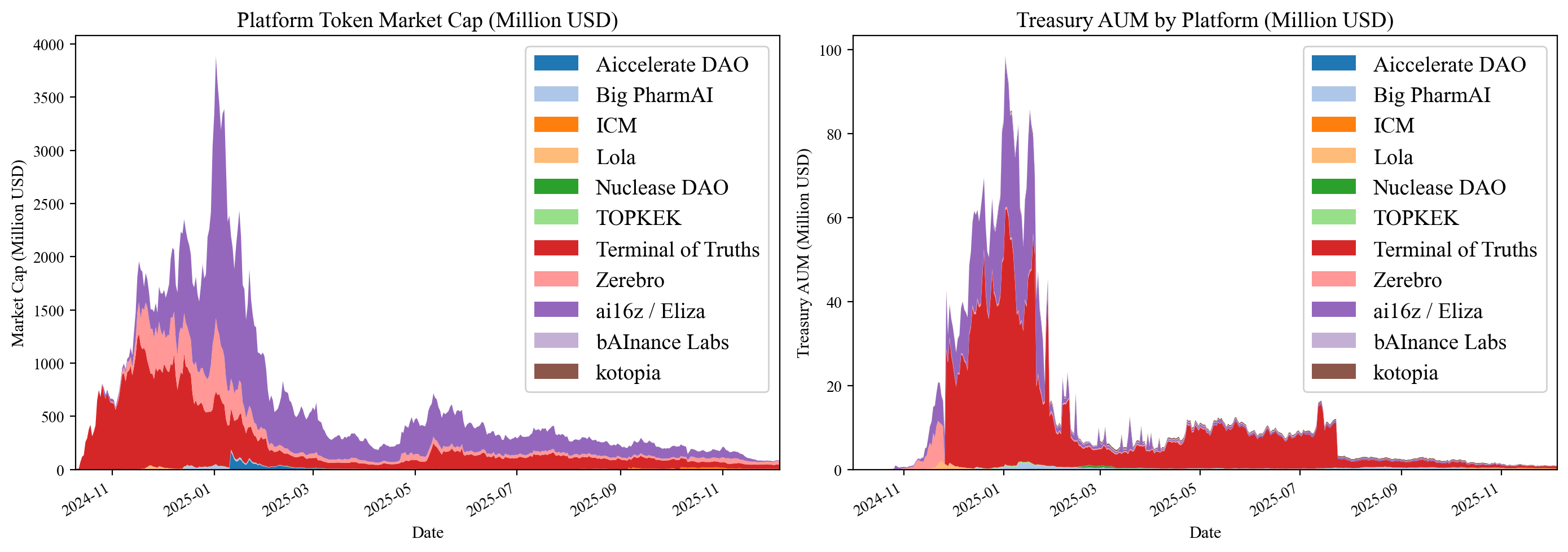}
    \caption{Platform token market capitalization (left) and treasury AUM (right) over time. ai16z/Eliza dominates both metrics.}
    \label{fig:mcap-aum-stacked}
\end{figure}

Platform valuations exhibit synchronized boom-bust dynamics (Figure~\ref{fig:mcap-aum-stacked}).

Market capitalizations peaked within a narrow window (late 2024 to early 2025), then declined sharply---agent tokens fell 93\% on average from all-time highs (range: 88--99\%) based on the market cap data. This clustering suggests valuations are driven by shared narrative cycles rather than idiosyncratic project fundamentals. ai16z/Eliza dominates the ecosystem, accounting for over 80\% of combined peak market capitalization.

\subsection{Treasury vs.\ User Performance}

\begin{figure}[t]
    \centering
    \includegraphics[width=0.95\linewidth]{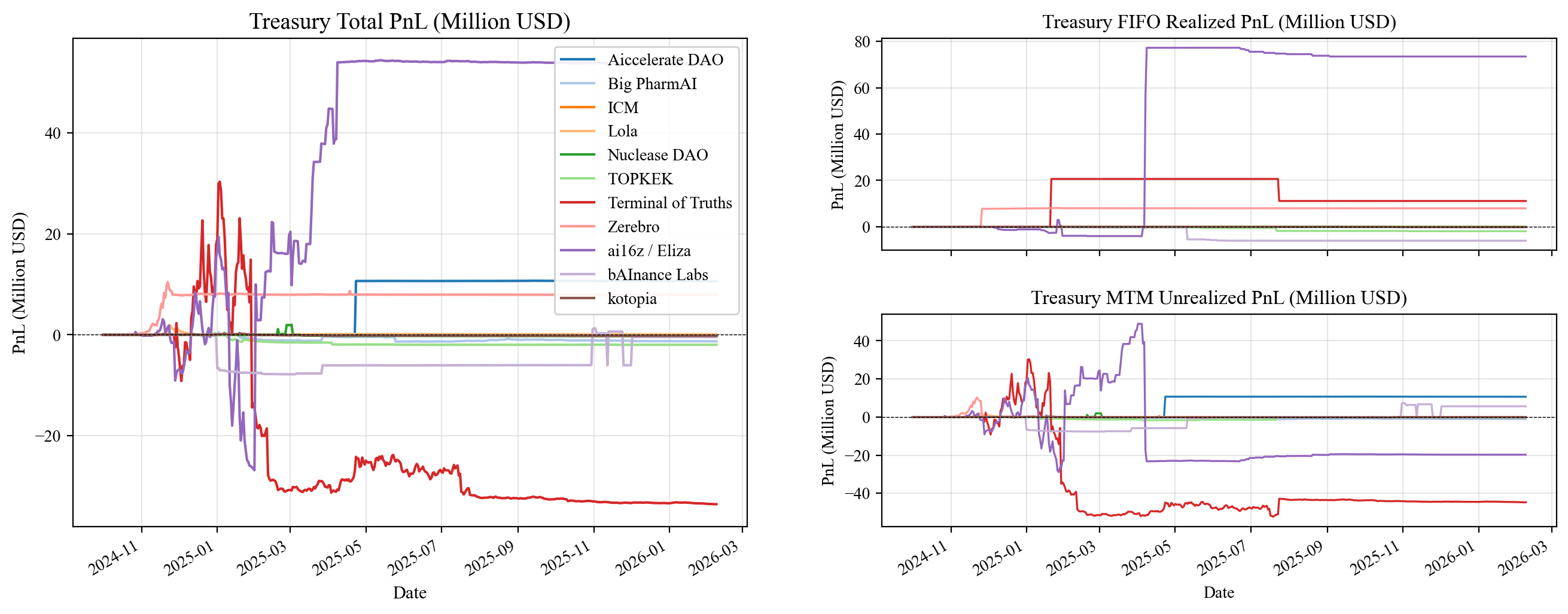}
    \caption{Treasury PnL by platform. Left: Total PnL (realized + unrealized). Right: breakdown into FIFO realized PnL (top) and MTM unrealized PnL (bottom).}
    \label{fig:treasury-pnl}
\end{figure}

\begin{figure}[t]
    \centering
    \includegraphics[width=0.95\linewidth]{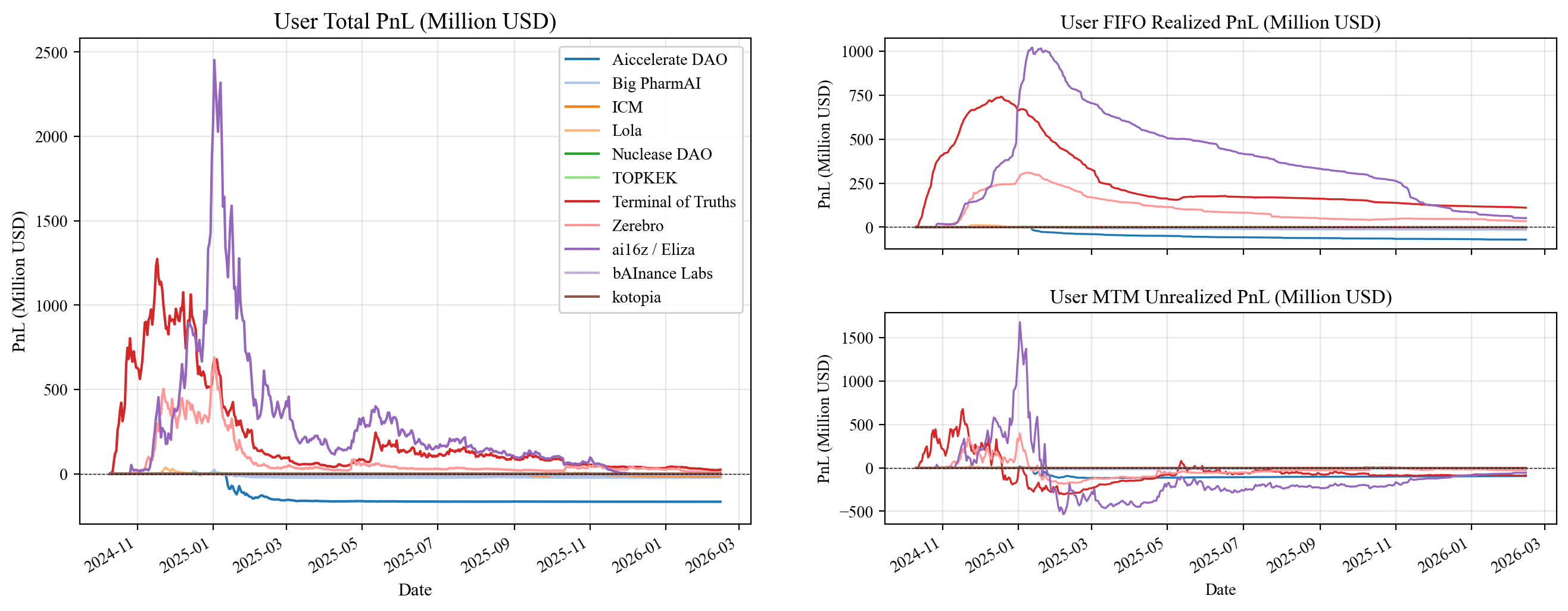}
    \caption{User PnL by platform. Left: Total PnL (realized + unrealized). Right: breakdown into FIFO realized PnL (top) and MTM unrealized PnL (bottom). Aggregate user PnL peaked at over \$2 billion before collapsing to net losses.}
    \label{fig:user-pnl}
\end{figure}

Treasury and user PnL exhibit different exposures across platforms rather than a uniform relationship (Figures~\ref{fig:treasury-pnl} and~\ref{fig:user-pnl}). Treasury outcomes vary substantially by platform: ai16z/Eliza, Aiccelerate DAO, and Zerebro retain positive treasury PnL in the latest snapshot, while several smaller treasuries are negative. User outcomes also vary, but aggregate user PnL has fallen from a peak of \$3.85B to a latest value of -\$191.7M across 925,323 wallets. These patterns suggest that treasury performance does not mechanically pass through to token-holder returns, but the relationship is heterogeneous rather than uniformly divergent.

This difference reflects the distinct assets held by treasuries and users. Treasuries hold baskets of portfolio tokens, often including early-stage or low-liquidity assets, while users primarily hold the platform's native token. As a result, treasury PnL reflects the performance of managed portfolio holdings, whereas user PnL reflects entry timing, token-market liquidity, and secondary-market demand for the platform token. Positive treasury PnL therefore need not imply positive user returns, and negative treasury PnL need not map one-to-one onto token-holder outcomes.

The boom-bust pattern is stark: at peak, aggregate user PnL across all platforms exceeded \$2.4 billion in paper gains (ai16z/Eliza alone reached \$2.45B). As of our latest snapshot by end of 2026 Feb, this has collapsed to a collective loss of \$191.7 million from 925,323 users across all platforms. Only two platforms---Truth Terminal (+\$24.1M) and Zerebro (+\$8.4M)---retain positive aggregate user PnL; all others show net losses. Note that user PnL accounts for both unrealized and realized profits as described in methodologies above.

\begin{figure}[t]
    \centering
    \includegraphics[width=0.95\linewidth]{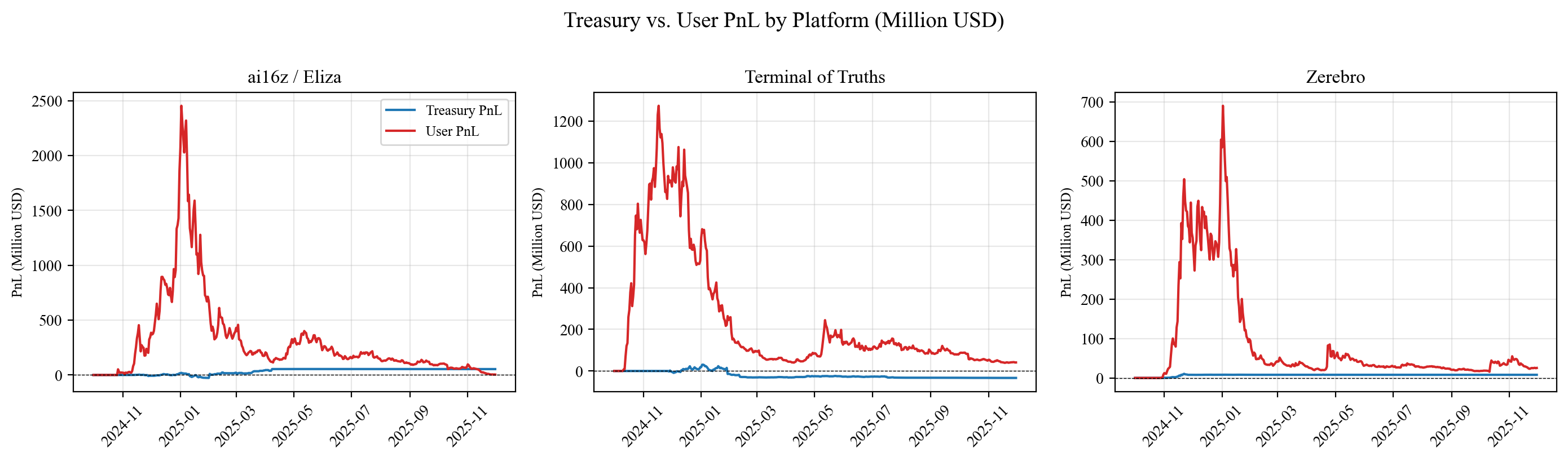}
    \caption{Treasury vs.\ user cumulative PnL by platform. Blue lines represent treasury PnL; red lines represent aggregate user PnL.}
    \label{fig:treasury-user-comparison}
\end{figure}

Figure~\ref{fig:treasury-user-comparison} illustrates the platform-level heterogeneity in treasury and user outcomes.

At the latest snapshot, aggregate treasury PnL remains positive at \$34.3M, while aggregate user PnL is negative at -\$191.7M. However, this aggregate contrast masks substantial platform-level variation. ai16z/Eliza shows strongly positive treasury PnL while user PnL is slightly negative; Aiccelerate DAO combines positive treasury PnL with large user losses; Zerebro is positive on both measures; and Truth Terminal is negative at the treasury level while users remain positive in aggregate. These cases indicate that the central empirical pattern is not a uniform treasury-user divergence, but rather the absence of a reliable pass-through mechanism from treasury performance to token-holder returns.

\subsection{User PnL Distribution}

User returns are heavily right-skewed (Table~\ref{tab:user-pnl-summary}). A small number of early participants capture most gains; the majority of later entrants realize losses. The single largest individual profit reached \$158.2 million (ai16z/Eliza), while median returns are negative or near-zero for all platforms, indicating that the typical participant loses money. Profitability rates range from 4--36\%, with the highest rates observed on Truth Terminal (35.8\%) and kotopia (27.5\%), and the lowest on Nuclease DAO (4.3\%) and Aiccelerate DAO (6.2\%).

Profit concentration is extreme: across all platforms, the top 1\% of profitable wallets (2,590 out of 259,016 winners) captured 81.4\% of all gains, totaling \$1.81 billion. At the platform level, concentration is similarly stark---for ai16z/Eliza, the top 1\% of winners (844 wallets) captured 83.6\% of profits; for Truth Terminal, 78.0\%; for Zerebro, 79.2\%. This pattern is consistent with early insiders and large allocators extracting the majority of value, while the remaining 62.2\% of participants (575,246 wallets) realized losses. Appendix~\ref{appendix:user-pnl-distribution} provides a time-series visualization of how user PnL distributions evolve across platforms.

\begin{table}[t]
\centering
\resizebox{\linewidth}{!}{%
\footnotesize
\begin{tabular}{lrrrrrrr}
\hline
\textbf{Platform} & \textbf{Users} & \textbf{Mean} & \textbf{Median} & \textbf{Min} & \textbf{Max} & \textbf{\% Prof.} & \textbf{\% Loss} \\
\hline
ai16z / Eliza & 323,320 & -\$14 & -\$3.93 & -\$75.5M & \$158.2M & 26.1\% & 66.6\% \\
Truth Terminal & 287,946 & \$84 & -\$0.01 & -\$96.8M & \$7.0M & 35.8\% & 53.8\% \\
Zerebro & 222,433 & \$38 & -\$0.19 & -\$43.9M & \$20.1M & 23.7\% & 65.8\% \\
Big PharmAI & 27,104 & -\$874 & -\$1.85 & -\$1.2M & \$114.8K & 26.7\% & 63.5\% \\
Lola & 21,038 & -\$141 & \$0.00 & -\$946.4K & \$345.2K & 25.9\% & 44.3\% \\
ICM & 20,745 & -\$723 & -\$21.05 & -\$813.2K & \$154.2K & 15.2\% & 76.2\% \\
Aiccelerate DAO & 9,982 & -\$16,614 & -\$29.69 & -\$56.7M & \$28.9K & 6.2\% & 78.1\% \\
bAInance Labs & 6,920 & -\$1,172 & -\$0.24 & -\$663.1K & \$10.3K & 9.5\% & 83.7\% \\
kotopia & 3,344 & -\$23 & \$0.00 & -\$90.1K & \$47.4K & 27.5\% & 37.2\% \\
TOPKEK & 2,051 & -\$28 & -\$0.10 & -\$133.8K & \$206.1K & 26.1\% & 56.5\% \\
Nuclease DAO & 440 & -\$9,061 & -\$10.07 & -\$3.2M & \$79 & 4.3\% & 75.2\% \\
\hline
\end{tabular}}
\caption{User PnL summary statistics by platform. Statistics are computed from a unified snapshot date with PnL forward-filled to reflect current valuations for all wallets. Users: unique wallets; Mean/Median/Min/Max: average, median, minimum, and maximum total PnL (USD); \% Prof./Loss: share of users with positive/negative PnL (remainder have exactly zero).}
\label{tab:user-pnl-summary}
\end{table}

\textbf{Developer perspectives on these results.} The distributional pattern in Table~\ref{tab:user-pnl-summary}---negative median PnL on every platform, with gains concentrated in a small fraction of wallets---is consistent with the capability gap our interview subjects independently described (Section~\ref{narrow-landscape}). Response from Eliza team characterized current LLM-based agents as unable to trade effectively without human-supplied insight, and described ElizaOS's intended trajectory as a ``trust marketplace'' in which humans contribute proprietary signals rather than as a source of autonomous trading skill. Virtuals' response similarly identified founder-side trading expertise---rather than the framework itself---as the differentiator between agents that perform and those that do not. Both teams thus attribute outcomes at the agent level to factors exogenous to the framework---the presence or absence of human trading expertise---which is consistent with our finding that user returns are dominated by a small set of profitable wallets while the median participant loses money.

\FloatBarrier
\subsection{Benchmark Comparison}

Agent tokens substantially underperformed a passive SOL buy-and-hold benchmark. During the study period (October 2024--November 2025), SOL experienced a 54\% peak-to-trough drawdown. Agent tokens declined 93\% on average---nearly twice the severity. Given that agent token peaks clustered within a narrow window, timing entry and exit was effectively impossible for typical users.

This underperformance, combined with negative median user PnL, highlights the difficulty of translating early agent-platform experimentation into reliable token-holder returns. Users face exposure not only to the underlying agent or treasury strategy, but also to token liquidity, market timing, narrative cycles, and secondary-market demand for the platform token.

The resulting return distribution is strongly right-skewed: a small share of wallets captured a disproportionate share of aggregate gains, while many participants realized losses. This creates a wealth-transfer pattern among token holders that resembles distributional dynamics documented in studies of Ponzi and pyramid schemes in blockchain contexts~\cite{bartoletti2020dissecting,kell2023forsage}. We emphasize, however, that this comparison is descriptive rather than accusatory: agent platforms do not necessarily promise fixed returns or intentionally rely on later entrants to pay earlier participants. Rather, the similarity lies in the observed payoff structure, where gains are highly concentrated and aggregate outcomes depend strongly on continued liquidity and secondary-market demand. This pattern is common in speculative token markets and other early-stage crypto ecosystems, where adoption waves, liquidity conditions, and timing can dominate fundamentals. We therefore interpret these outcomes as evidence of immature market structure and weak pass-through from agent activity to token-holder returns, rather than as evidence that current platforms intentionally engineer losses for users.

\section{Conditions for Market Maturity} \label{sec:maturity}

Our empirical findings characterize an early-stage market whose current deployments remain short of the requirements for robust autonomous investment systems. These limitations are not inevitable; rather, they reflect a gap between the capabilities DeFi investment agents would need to operate reliably and the infrastructure, liquidity, verification mechanisms, and incentive designs currently available. We outline three areas of development before proposing a maturity framework for evaluating progress toward production-grade autonomous trading systems.

\subsection{Prerequisites for Maturity}
The underperformance we document stems from three structural problems. First, DeFi markets lack the infrastructure that makes algorithmic trading viable: shallow liquidity means agents move prices with their own trades (our agents concentrated in tokens under \$10M market cap), fragmented execution across chains prevents reliable backtesting, and MEV exploitation~\cite{daian2020flash} introduces unpredictable costs. Second, the memecoin focus we observed creates pure reflexivity: prices driven by attention rather than fundamentals~\cite{Conlon2025MemecoinContagion} means patterns vanish when viral traction fades, and our finding of synchronized peaks across agents confirms crowded trades that self-destruct. Third, the extreme MC-to-AUM ratios we documented (exceeding 10,000$\times$ versus below 1$\times$ for established protocols) reflect the CoinAlg Bind's prediction~\cite{fabrega2026coinalg}: platforms earn fees on volume regardless of user outcomes, creating misalignment that our negative median user returns confirm.

\subsection{Proposed Maturity Framework}
Based on these findings, we propose three \emph{maturity criteria} for DeFi investment agents. These are distinct from the inclusion criteria of Section~\ref{broad-landscape}, which scoped the dataset; here we assess the agents that survived that filter:

\mypara{Criterion 1: Verifiable Autonomous Execution.}
A mature DeFi investment agent should make its execution process externally verifiable. This requires more than publishing wallet addresses: observers should be able to determine whether on-chain transactions were initiated by the declared agent logic, a human operator, or another off-chain process. Future systems could support this through auditable strategy configurations, signed execution traces, trusted execution environments, zero-knowledge attestations, or standardized logs linking decisions to trades. Such mechanisms would allow permissionless agent platforms to support a wide range of developer strategies while still giving users a basis for evaluating autonomy.

\mypara{Criterion 2: Sustained Risk-Adjusted Performance.}
A mature agent should report performance in a way that distinguishes durable trading skill from exposure to a favorable market cycle. Performance evaluation should therefore cover multiple market regimes, compare returns against passive benchmarks such as SOL or ETH, and include standard risk measures such as Sharpe ratio, maximum drawdown, turnover, liquidity-adjusted returns, and realized versus unrealized gains. For tokenized agents, evaluation should also separate treasury-level performance from token-holder outcomes, since the assets held by the treasury and by users can differ substantially. This would make it clearer whether value is generated by the agent's trading process, by appreciation of the platform token, or by broader market movements.

\mypara{Criterion 3: Transparent Stakeholder Alignment.}
A mature platform should make the relationship between agent performance, platform revenue, treasury value, and token-holder returns explicit. Potential mechanisms include NAV-based redemption, pro-rata distributions, systematic buybacks, performance fees with high-water marks, or governance processes that condition fees and treasury actions on realized outcomes. The key requirement is not a single prescribed design, but transparency: users should be able to understand how value created by an agent, if any, flows to the stakeholders who provide capital or bear token-market risk. This is especially important in permissionless markets, where platform infrastructure may enable many independently deployed agents with different strategies and incentive structures.

Together, these criteria provide a forward-looking framework for evaluating progress in DeFi investment agents. The first generation of platforms has enabled rapid experimentation in a permissionless setting, but the market has not yet converged on common standards for verifying autonomy, measuring performance, or aligning stakeholder incentives. We therefore view the observed limitations as signs of an immature market structure rather than as evidence that the underlying platform model is inherently flawed. Progress will likely require both stronger market infrastructure---deeper liquidity, reliable execution, and less adversarial ordering---and clearer incentive designs that connect agent activity to user-facing value.

\section{Conclusion}
\label{sec:conclusion}

We survey over 1,900 AI-tagged crypto projects, filter to investment-focused agents, and curate a set of 10 representative projects spanning strategy and observability dimensions. We then conduct a deep-dive architectural analysis of the two dominant frameworks---ElizaOS and Virtuals Protocol---and a quantitative on-chain performance analysis of 11 Solana-based agent treasuries (covering 925,323 token holders) for which trading activity is publicly attributable, complemented by developer interviews.

Four findings recur across every level of analysis. First, most agents do not autonomously execute trades, and developer interviews confirm that even on platforms with over 17,000 agent launches, the vast majority implement ``basic API integrations'' rather than autonomous trading. Second, agent treasuries retain over \$30M in paper gains while their token holders collectively lost \$191.7M---aggregate user gains peaked at \$2.4B before collapsing to net losses. Third, the top 1\% of profitable wallets captured 81.4\% of all gains (\$1.81B), while median returns are negative on every platform. Fourth, token valuations are decoupled from fundamentals, with MC-to-AUM ratios exceeding 10,000$\times$ versus below 1$\times$ for established DeFi protocols, and tokens declining 93\% on average from all-time highs.

We formalize these observations into a maturity framework---\emph{verifiable autonomous execution}, \emph{sustained risk-adjusted performance}, and \emph{transparent stakeholder alignment}. Rather than treating these criteria as a pass/fail judgment on first-generation systems, we use them to identify the standards that future DeFi investment agents would need to satisfy to function as robust autonomous investment vehicles.

\textbf{Limitations.} Our analysis is bounded by observable wallet addresses, daily balance snapshots rather than individual transaction traces, a single market cycle (October 2024--November 2025), and the difficulty of distinguishing autonomous execution from human-in-the-loop operation using public on-chain data alone.

\textbf{Future Work.} Mechanisms for \emph{verifiable autonomy}---such as TEEs or zero-knowledge proofs that trading decisions originate from declared algorithms---could help address the gap between observed transactions and verifiable agent execution. The high MC-to-AUM ratios we observe motivate theoretical analysis of equilibrium pricing for tokenized on-chain investment vehicles, building on the CoinAlg Bind framework's~\cite{fabrega2026coinalg} fairness--arbitrage tradeoff. Finally, longitudinal studies across multiple market cycles would clarify which dynamics are specific to first-generation DeFi investment agents and which persist as the market matures.

\textbf{Acknowledgment.} Thanks to JW Seo for the discussion on data methodology, Eliza and Virtuals team for the interviews, James Austgen, Andres Fabrega, and Ari Juels for their comments and reviews.


\ifACM \bibliographystyle{ACM-Reference-Format.bst} \fi
\ifUSENIX \bibliographystyle{plain} \fi
\ifIEEE \bibliographystyle{plain} \fi
\ifLNCS \bibliographystyle{Conferences/LNCS/splncs04} \fi

\iffull \else \bibliography{references} \fi

@inproceedings{daian2020flash,
  title={Flash boys 2.0: Frontrunning in decentralized exchanges, miner extractable value, and consensus instability},
  author={Daian, Philip and Goldfeder, Steven and Kell, Tyler and Li, Yunqi and Zhao, Xueyuan and Bentov, Iddo and Breidenbach, Lorenz and Juels, Ari},
  booktitle={IEEE S\&P},
  year={2020},
}

@inproceedings{qin2021attacking,
  title={Attacking the defi ecosystem with flash loans for fun and profit},
  author={Qin, Kaihua and Zhou, Liyi and Livshits, Benjamin and Gervais, Arthur},
  booktitle={International conference on financial cryptography and data security},
  pages={3--32},
  year={2021},
  organization={Springer}
}

@inproceedings{qin2021empirical,
  title={An empirical study of defi liquidations: Incentives, risks, and instabilities},
  author={Qin, Kaihua and Zhou, Liyi and Gamito, Pablo and Jovanovic, Philipp and Gervais, Arthur},
  booktitle={Proceedings of the 21st ACM internet measurement conference},
  pages={336--350},
  year={2021}
}

@misc{CoinMarketCapSolanaEcosystem2025,
  title        = {Top Solana Ecosystem Tokens by Market Capitalization},
  howpublished = {\url{https://coinmarketcap.com/view/solana-ecosystem/}},
  note         = {Accessed: 2025-06-25},
  author       = {{CoinMarketCap}},
  year         = {2025},
  month        = jun,
}

@misc{walters2025elizaweb3friendlyai,
      title={Eliza: A Web3 friendly AI Agent Operating System}, 
      author={Shaw Walters and Sam Gao and Shakker Nerd and Feng Da and Warren Williams and Ting-Chien Meng and Amie Chow and Hunter Han and Frank He and Allen Zhang and Ming Wu and Timothy Shen and Maxwell Hu and Jerry Yan},
      year={2025},
      eprint={2501.06781},
      archivePrefix={arXiv},
      primaryClass={cs.AI},
      url={https://arxiv.org/abs/2501.06781}, 
}

@misc{crypto_elizaos,
  title        = {What is ElizaOS?},
  howpublished = {\url{https://crypto.com/es/university/what-is-elizaos}},
  note         = {Accessed: 2025-09-08},
  author       = {{Crypto.com University}},
  year         = {2025}
}

@misc{decrypt_ai16z_launch,
  title        = {Meet ai16z, an {AI}-Based Investment {DAO} Inspired by Marc Andreessen},
  author       = {{Decrypt}},
  year         = {2024},
  howpublished = {\url{https://decrypt.co/295717/meet-ai16z-dao-an-ai-based-investment-project-that-aims-to-upend-silicon-valley}},
  note         = {Accessed: 2026-05-05}
}

@misc{decrypt_elizaos_rename,
  title        = {{AI} {DAO} ai16z Becomes ElizaOS Amid Branding Confusion Concerns},
  author       = {{Decrypt}},
  year         = {2025},
  howpublished = {\url{https://decrypt.co/303244/ai-dao-ai16z-becomes-elizaos-amid-branding-confusion-concerns}},
  note         = {Accessed: 2026-05-05}
}

@misc{coinbase_ai16z,
  title        = {AI16z Price},
  howpublished = {\url{https://www.coinbase.com/price/ai16z}},
  note         = {Accessed: 2025-09-08},
  author       = {{Coinbase}},
  year         = {2025}
}

@misc{VirtualsProtocolWhitepaper2025,
  title        = {Virtuals Protocol Whitepaper},
  howpublished = {\url{https://whitepaper.virtuals.io/#background}},
  author       = {{Virtuals Protocol}},
  year         = {2025},
  note         = {Accessed: 2025-06-25},
}

@misc{coinmarketcap,
  title        = {CoinMarketCap},
  author       = {{CoinMarketCap}},
  year         = {2025},
  url          = {https://coinmarketcap.com},
  note         = {Accessed: 2025-09-06}
}

@article{ji2017robots,
  title={{Are Robots Good Fiduciaries? Regulating Robo-Advisors Under the Investment Advisers Act of 1940}},
  author={Ji, Megan},
  journal={Columbia Law Review},
  year={2017},
}

@inproceedings{Sharma2024DAOs,
  author       = {Tanusree Sharma and Yujin Potter and Kornrapat Pongmala and Henry Wang and Andrew Miller and Dawn Song and Yang Wang},
  title        = {Unpacking How Decentralized Autonomous Organizations (DAOs) Work in Practice},
  booktitle    = {2024 IEEE International Conference on Blockchain and Cryptocurrency (ICBC 2024)},
  pages        = {416--424},
  year         = {2024},
  doi          = {10.1109/ICBC59979.2024.10634404},
  publisher    = {Institute of Electrical and Electronics Engineers (IEEE)},
  address      = {Dublin, Ireland}
}

@article{Conlon2025MemecoinContagion,
  author       = {T. Conlon and S. Corbet and others},
  title        = {Memecoin contagion: Irrationality, illicit behaviour, and contagion in rapid memecoin growth},
  journal      = {Journal of Behavioral and Experimental Finance},
  year         = {2025},
  volume       = {50},
  pages        = {101518},
  doi          = {10.1016/j.jbef.2025.101518},
  url          = {https://doi.org/10.1016/j.jbef.2025.101518}
}

@misc{Bellan2024TruthTerminal,
  author       = {Rebecca Bellan},
  title        = {The promise and warning of {Truth Terminal}, the AI bot that secured \$50,000 in bitcoin from Marc Andreessen},
  year         = {2024},
  month        = dec,
  howpublished = {\url{https://techcrunch.com/2024/12/19/the-promise-and-warning-of-truth-terminal-the-ai-bot-that-secured-50000-in-bitcoin-from-marc-andreessen/}},
  note         = {TechCrunch. Accessed: 2025-11-21}
}

@misc{proj:Axal,
  title        = {Axal: Quick Start Guide},
  author       = {Axal},
  year         = {2025},
  howpublished = {\url{https://docs.axal.com/getting-started/quick-start/}},
  note         = {Accessed: 2025-11-21}
}

@misc{proj:GizaProtocol,
  title        = {Giza Protocol Documentation},
  author       = {Giza},
  year         = {2025},
  howpublished = {\url{https://docs.gizaprotocol.ai/}},
  note         = {Accessed: 2025-11-21}
}

@misc{proj:TrueNorthXYZ,
  title        = {TrueNorth – Crypto’s First AI Discovery Engine},
  author       = {TrueNorth},
  year         = {2025},
  howpublished = {\url{https://www.true-north.xyz/}},
  note         = {Accessed: 2025-11-21}
}

@misc{proj:AskSurfAI,
  title        = {AskSurf.ai – Crypto’s Ultimate AI},
  author       = {AskSurf},
  year         = {2025},
  howpublished = {\url{https://asksurf.ai/}},
  note         = {Accessed: 2025-11-21}
}

@misc{proj:AIXBT,
  title        = {AIXBT.tech – AI-Driven Crypto Market Intelligence},
  author       = {AIXBT},
  year         = {2025},
  howpublished = {\url{https://aixbt.tech/}},
  note         = {Accessed: 2025-11-21}
}

@misc{proj:TruthTerminal,
  title        = {TruthTerminal.wiki – The Truth Terminal Project},
  author       = {Truth Terminal},
  year         = {2025},
  howpublished = {\url{https://truthterminal.wiki/}},
  note         = {Accessed: 2025-11-21}
}

@misc{proj:AXR_AIXVC,
  title        = {AXR / Axelrod by AIxVC: Agentic Quant Fund Platform},
  author       = {AIxVC / Axelrod},
  year         = {2025},
  howpublished = {\url{https://axr.aixvc.io/}},
  note         = {Accessed: 2025-11-21}
}

@misc{proj:ElizaOS,
  title        = {ElizaOS.ai – Agentic Operating System for AI Agents},
  author       = {ElizaOS},
  year         = {2025},
  howpublished = {\url{https://www.elizaos.ai/}},
  note         = {Accessed: 2025-11-21}
}

@misc{proj:NoF1AI,
  title        = {NoF1.ai – Project Homepage},
  author       = {NoF1},
  year         = {2025},
  howpublished = {\url{https://nof1.ai/}},
  note         = {Accessed: 2025-11-21}
}

@misc{proj:NumeraiCrypto,
  title        = {Numerai Crypto – Overview},
  author       = {Numerai},
  year         = {2025},
  howpublished = {\url{https://docs.numer.ai/numerai-crypto/crypto-overview}},
  note         = {Accessed: 2025-11-21}
}

@misc{PanteraDATboard,
  title        = {Pantera DATboard},
  howpublished = {\url{https://datboard.panteraresearchlab.xyz/}},
  note         = {Accessed: 2025-12-03},
  organization = {Pantera Research Lab}
}

@article{fabrega2026coinalg,
  title={The {CoinAlg} Bind: Profitability-Fairness Tradeoffs in Collective Investment Algorithms},
  author={F{\'{a}}brega, Andr{\'{e}}s and Austgen, James and Breckenridge, Samuel and Yu, Jay and Zhao, Amy and Allen, Sarah and Saraf, Aditya and Juels, Ari},
  journal={arXiv preprint arXiv:2601.00523},
  year={2026}
}

@inproceedings{xu2019anatomy,
  title={The Anatomy of a Cryptocurrency Pump-and-Dump Scheme},
  author={Xu, Jiahua and Livshits, Benjamin},
  booktitle={Proceedings of the 28th USENIX Security Symposium},
  pages={1609--1625},
  year={2019}
}

@misc{elizaos_plugin_solana,
  title        = {plugin-solana: Core {Solana} blockchain plugin for {ElizaOS}},
  author       = {{ElizaOS}},
  year         = {2025},
  howpublished = {\url{https://github.com/elizaos-plugins/plugin-solana}},
  note         = {Accessed: 2025-11-30}
}

@misc{elizaos_plugin_hyperliquid,
  title        = {plugin-hyperliquid: {Hyperliquid} {DEX} integration for {ElizaOS}},
  author       = {{ElizaOS}},
  year         = {2025},
  howpublished = {\url{https://github.com/elizaos-plugins/plugin-hyperliquid}},
  note         = {Accessed: 2025-11-30}
}

@misc{elizaos_plugin_registry,
  title        = {ElizaOS Plugin Registry},
  author       = {{ElizaOS}},
  year         = {2025},
  howpublished = {\url{https://github.com/elizaos-plugins/registry}},
  note         = {Accessed: 2025-11-30}
}

@misc{elizaos_docs,
  title        = {ElizaOS Developer Documentation: Plugin Components},
  author       = {{ElizaOS}},
  year         = {2025},
  howpublished = {\url{https://docs.elizaos.ai/plugins/components}},
  note         = {Accessed: 2025-11-30}
}

@misc{virtuals_game_framework,
  title        = {{GAME} Framework: Generative Autonomous Multimodal Entities},
  author       = {{Virtuals Protocol}},
  year         = {2025},
  howpublished = {\url{https://whitepaper.virtuals.io/builders-hub/game-framework}},
  note         = {Accessed: 2025-11-30}
}

@misc{virtuals_acp_deepdive,
  title        = {Agent Commerce Protocol: Technical Deep Dive},
  author       = {{Virtuals Protocol}},
  year         = {2025},
  howpublished = {\url{https://whitepaper.virtuals.io/about-virtuals/agent-commerce-protocol-acp/technical-deep-dive}},
  note         = {Accessed: 2025-11-30}
}

@misc{virtuals_protocol_contracts,
  title        = {Virtuals Protocol Smart Contracts},
  author       = {{Virtuals Protocol}},
  year         = {2025},
  howpublished = {\url{https://github.com/Virtual-Protocol/protocol-contracts}},
  note         = {Accessed: 2025-11-30}
}

@misc{virtuals_game_sdk,
  title        = {{GAME} {Python} {SDK}},
  author       = {{Virtuals Protocol}},
  year         = {2025},
  howpublished = {\url{https://github.com/game-by-virtuals/game-python}},
  note         = {Accessed: 2025-11-30}
}

@article{chen2025stockbench,
  title={{StockBench}: Can {LLM} Agents Trade Stocks Profitably In Real-world Markets?},
  author={Chen, Yanxu and Yao, Zijun and Liu, Yantao and Ye, Jin and Yu, Jianing and Hou, Lei and Li, Juanzi},
  journal={arXiv preprint arXiv:2510.02209},
  year={2025}
}

@article{fan2025aitrader,
  title={{AI-Trader}: Benchmarking Autonomous Agents in Real-Time Financial Markets},
  author={Fan, Tianyu and Yang, Yuhao and Jiang, Yangqin and Zhang, Yifei and Chen, Yuxuan and Huang, Chao},
  journal={arXiv preprint arXiv:2512.10971},
  year={2025}
}

@article{qian2025agentstrade,
  title={When Agents Trade: Live Multi-Market Trading Benchmark for {LLM} Agents},
  author={Qian, Lingfei and Peng, Xueqing and Wang, Yan and Zhang, Vincent Jim and He, Huan and Smith, Hanley and Han, Yi and He, Yueru and Li, Haohang and Cao, Yupeng and Yu, Yangyang and Lopez-Lira, Alejandro and Lu, Peng and Nie, Jian-Yun and Xiong, Guojun and Huang, Jimin and Ananiadou, Sophia},
  journal={arXiv preprint arXiv:2510.11695},
  year={2025}
}

@article{yang2023fingpt,
  title={{FinGPT}: Open-Source Financial Large Language Models},
  author={Yang, Hongyang and Liu, Xiao-Yang and Wang, Christina Dan},
  journal={arXiv preprint arXiv:2306.06031},
  year={2023}
}

@article{luo2025llmcrypto,
  title={{LLM}-Powered Multi-Agent System for Automated Crypto Portfolio Management},
  author={Luo, Yichen and Feng, Yebo and Xu, Jiahua and Tasca, Paolo and Liu, Yang},
  journal={arXiv preprint arXiv:2501.00826},
  year={2025}
}

@misc{coingecko_ai_agents,
  title        = {Top {AI} Agents Coins by Market Cap},
  author       = {{CoinGecko}},
  year         = {2025},
  howpublished = {\url{https://www.coingecko.com/en/categories/ai-agents}},
  note         = {Accessed: 2025-11-30}
}

@misc{messari_fundraising,
  title        = {Crypto Fundraising Database},
  author       = {{Messari}},
  year         = {2025},
  howpublished = {\url{https://messari.io/fundraising-data}},
  note         = {Accessed: 2025-11-30}
}

@misc{dune,
  title        = {{Dune}: Onchain Data Platform},
  author       = {{Dune Analytics}},
  year         = {2025},
  howpublished = {\url{https://dune.com/}},
  note         = {Accessed: 2025-02-14}
}

@misc{dune:platformpnl,
  title        = {Platform Treasury {PnL} with {FIFO} Realized and {MTM} Unrealized},
  author       = {{Dune Analytics}},
  year         = {2025},
  howpublished = {\url{https://dune.com/queries/6697062}},
  note         = {Accessed: 2025-02-14}
}

@misc{dune:userbalance,
  title        = {User Daily Balance Snapshots for Agent Platform Tokens},
  author       = {{Dune Analytics}},
  year         = {2025},
  howpublished = {\url{https://dune.com/queries/6697157}},
  note         = {Accessed: 2025-02-14}
}

@misc{dune:tokenprice,
  title        = {Token Daily Price Snapshots},
  author       = {{Dune Analytics}},
  year         = {2025},
  howpublished = {\url{https://dune.com/queries/6697297}},
  note         = {Accessed: 2025-02-14}
}

@inproceedings{bartoletti2020dissecting,
  title={Dissecting Ponzi schemes on Ethereum: Identification, analysis, and impact},
  author={Bartoletti, Massimo and Carta, Salvatore and Cimoli, Tiziana and Saia, Roberto},
  booktitle={Future Generation Computer Systems},
  volume={102},
  pages={259--277},
  year={2020},
  publisher={Elsevier}
}

@inproceedings{kell2023forsage,
  title={Forsage: Anatomy of a smart-contract pyramid scheme},
  author={Kell, Tyler and Cr\'{e}pault, Haaroon and Hua, Yan and Liu, Siyi and Sun, Xiaofan and Mudgal, Chaitra and Juels, Ari},
  booktitle={Financial Cryptography and Data Security (FC)},
  year={2023}
}

\ifSP \appendices
    \iffull
    \else
    \input{Sections/A-other-applications}
    \input{Sections/A-liveness-details}
    \input{Sections/A-complete-knowledge-details}

    \fi
\else
  \ifLNCS
    \appendix
    \section{User PnL Distribution Over Time}
\label{appendix:user-pnl-distribution}

\begin{figure}[H]
    \centering
    \includegraphics[width=\linewidth]{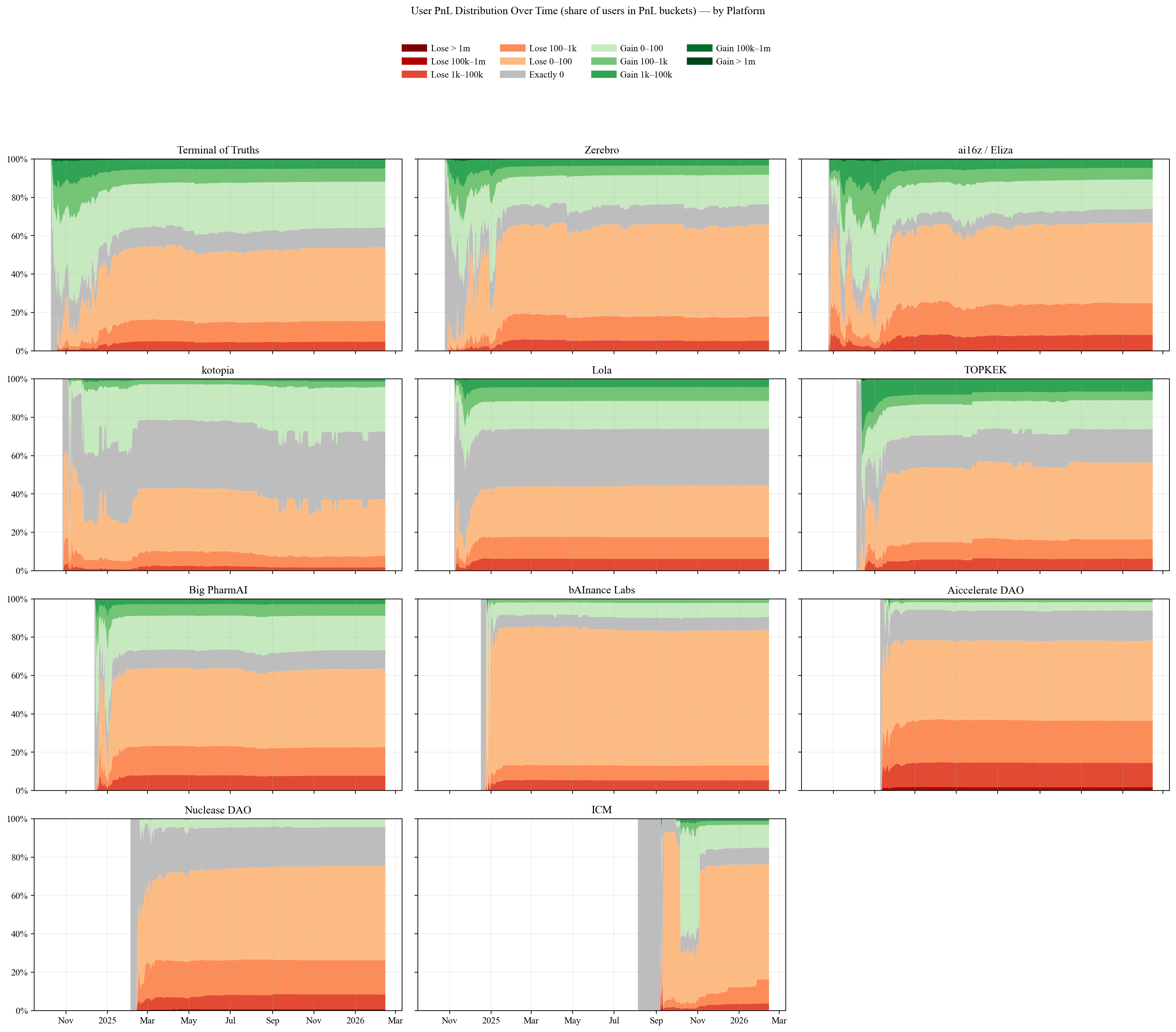}
    \caption{User PnL distribution over time by platform. Each subplot shows the share of users in different PnL buckets (from large losses in red to large gains in green, with exact zero in gray). PnL values are forward-filled to the latest snapshot date for all wallets.}
    \label{fig:user-pnl-dist-time}
\end{figure}

  \else
    \appendix
    \section{User PnL Distribution Over Time}
\label{appendix:user-pnl-distribution}

\begin{figure}[H]
    \centering
    \includegraphics[width=\linewidth]{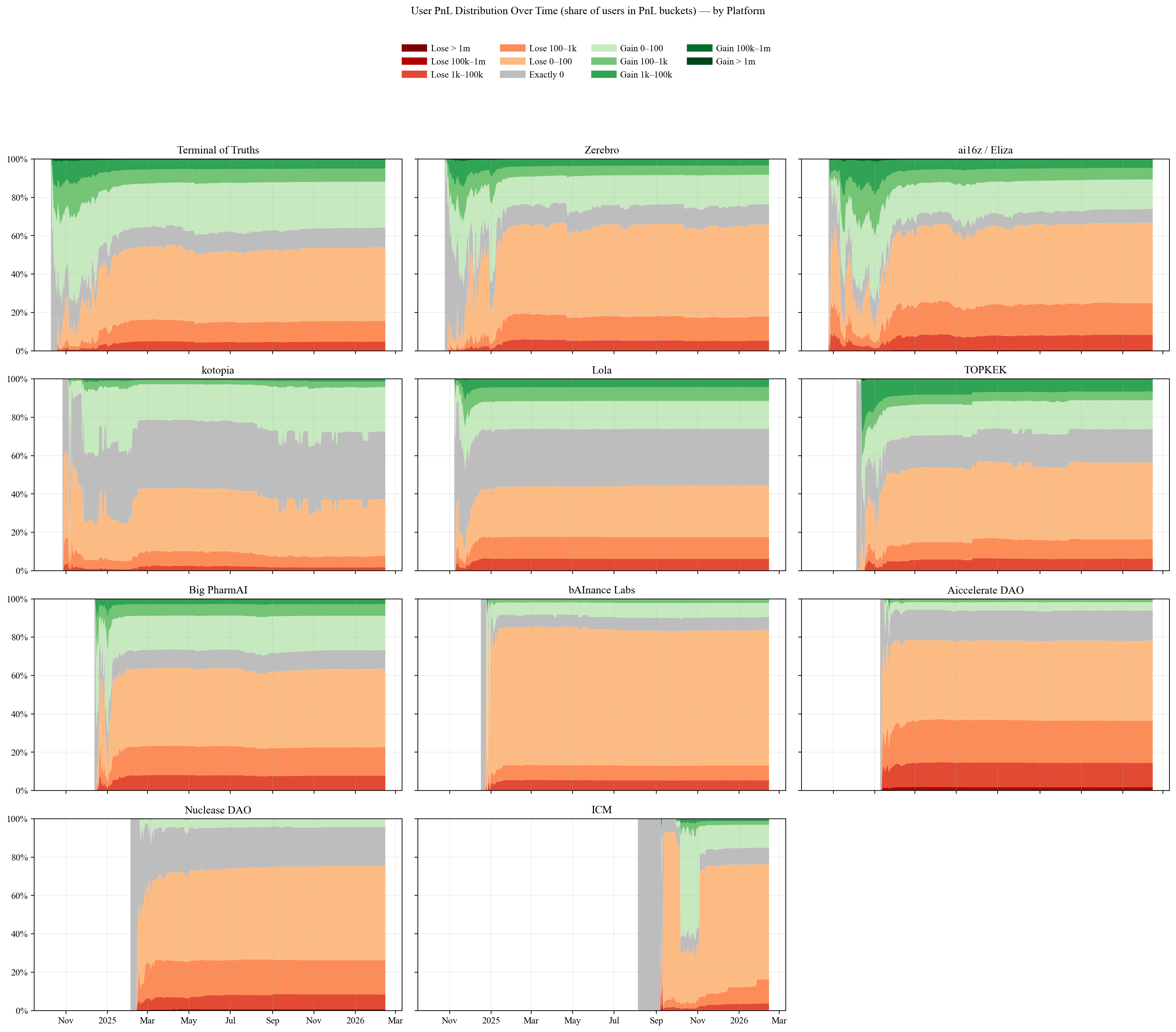}
    \caption{User PnL distribution over time by platform. Each subplot shows the share of users in different PnL buckets (from large losses in red to large gains in green, with exact zero in gray). PnL values are forward-filled to the latest snapshot date for all wallets.}
    \label{fig:user-pnl-dist-time}
\end{figure}

  \fi
\appendix \fi
\end{document}